%% file: main.tex
\documentclass[runningheads]{llncs}
\usepackage{graphicx}

\usepackage{tikz}
\usepackage{comment}
\usepackage{amsmath,amssymb} 
\usepackage{color}
\usepackage{subcaption}
\usepackage{multirow}
\usepackage{booktabs}
\usepackage[accsupp]{axessibility}  

\begin{document}

\pagestyle{headings}
\mainmatter
\def\ECCVSubNumber{6193}  

\title{Unsupervised Segmentation in Real-World Images via Spelke Object Inference}


%
\author{
Honglin Chen \inst{1} \and
Rahul Venkatesh \inst{1} \and
Yoni Friedman \inst{4} \and
Jiajun Wu \inst{1} \and \\
Joshua B. Tenenbaum \inst{4}\index{Tenenbaum, Joshua B.} \and
Daniel L. K. Yamins \inst{1,2,3,**}\index{Yamins, Daniel L. K.} \and
Daniel M. Bear \inst{2,3,**}\index{Bear, Daniel M.}
}
\authorrunning{Chen et al.}
\titlerunning{Unsupervised Segmentation via Spelke Object Inference}
%

\institute{Department of Computer Science, Stanford \and 
Department of Psychology, Stanford \and
Wu~Tsai~Neurosciences~Institute, Stanford \and
Department of Brain and Cognitive Sciences and CBMM, MIT\\
\email{\{honglinc,dbear\}@stanford.edu}\\
$^{**}=$ Equal contribution} 

\maketitle

\begin{abstract}
Self-supervised, category-agnostic segmentation of real-world images is a challenging open problem in computer vision.  
Here, we show how to learn static grouping priors from motion self-supervision by building on the cognitive science concept of a Spelke Object: a set of physical stuff that moves together.  
We introduce the Excitatory-Inhibitory Segment Extraction Network (EISEN), which learns to extract pairwise affinity graphs for static scenes from motion-based training signals.
EISEN then produces segments from affinities using a novel graph propagation and competition network. 
During training, objects that undergo correlated motion (such as robot arms and the objects they move) are decoupled by a bootstrapping process: EISEN explains away the motion of objects it has already learned to segment.
We show that EISEN achieves a substantial improvement in the state of the art for self-supervised image segmentation on challenging synthetic and real-world robotics datasets.
\end{abstract}

\input{Sections/introduction_CR}

\input{Sections/related_work_CR}
\input{Sections/methods_CR}
\input{Sections/results_CR}

\input{Sections/conclusion}

\clearpage
%
%
\bibliographystyle{splncs04}

\input{main.bbl}
\newpage
\appendix

\input{Sections/supplement}

\newpage
\end{document}

%% file: Sections/introduction_CR.tex
\section{Introduction}
Most approaches to image segmentation rely heavily on supervised data that is challenging to obtain and are largely trained in a category-specific way~\cite{shi2000normalized,girshick2015fast,he2017mask,cheng2020panoptic}. Thus, even state of the art segmentation networks struggle with recognizing untrained object categories and complex configurations~\cite{follmann2018mvtec}. A self-supervised, category-agnostic segmentation algorithm would be of great value.  

But how can a learning signal for such an algorithm be obtained? The cognitive science of perception in babies provides a clue, via the concept of a \emph{Spelke object}~\cite{spelke1990principles}: a collection of physical stuff that moves together under the application of everyday physical actions.\footnotetext[2]{More formally, two pieces of stuff are considered to be in the same Spelke object if and only if, under the application of any sequence of actions that causes sustained motion of one of the pieces of stuff, the magnitude of the motion that the other piece of stuff experiences relative to the first piece is approximately zero compared to the magnitude of overall motion. Natural action groups arise from the set of all force applications exertable by specific physical actuator, such as (e.g.) a pair of human hands or a robotic gripper.} Perception of Spelke objects is category-agnostic and acquired by infants without supervision~\cite{spelke1990principles}. In this work we build a neural network that learns from motion signals to segment Spelke objects in still images (Fig. \ref{fig:headliner}). To achieve this goal, we make two basic innovations.
\input{Figures/fig1}

\input{Figures/fig2}
First, we design a pairwise affinity-based grouping architecture that is optimized for learning from motion signals. Most modern segmentation networks are based on pixelwise background-foreground categorization ~\cite{he2017mask,cheng2020panoptic}. However, Spelke objects are fundamentally relational, in that they represent whether \textit{pairs} of scene elements are likely to move together. Moreover, this physical connectivity must be learned from real-world video data in which motion is comparatively sparse, as only one or a few Spelke objects is typically moving at a time (Fig. \ref{fig:challenges}, top). Standard pixelwise classification problems that attempt to approximate these pairwise statistics (such as the ``Spelke-object-or-not'' task) induce large numbers of false negatives for temporarily non-moving objects. Directly learning pairwise affinities avoids these problems.

To convert affinities into actual segmentations, we implement a fully differentiable grouping network inspired by the neuroscience concepts of recurrent label propagation and border ownership cells~\cite{roelfsema2006cortical,zhou2000coding}. This network consists of (i) a quasi-local, recurrent affinity Propagation step that creates (soft) segment identities across pixels in an image and (ii) a winner-take-all Competition step that assigns a unique group label to each segment.  We find through ablation studies that this specific grouping mechanism yields high-quality segments from affinities.  

A second innovation is an iterative scheme for network training. In real-world video, most objects are inanimate, and thus only seen in motion when caused to move by some other animate object, such as a human hand or robotic gripper (Fig \ref{fig:challenges}, bottom). This correlated motion must therefore be dissociated to learn to segment the mover from the moved object. Cognitive science again gives a clue to how this may be done: babies first learn to localize hands and arms, then later come to understand external objects~\cite{ullman2012simple}. We implement this concept as a confidence-thresholded bootstrapping procedure: motion signals that are already well-segmented by one iteration of network training are explained away, leaving unexplained motions to be treated as independent sources for supervising the next network iteration. For example, in natural video datasets with robotic grippers, the gripper arm will naturally arise as a high-confidence segment first, allowing for the object in the gripper to be recognized as a separate object via explaining-away.
The outputs of this explaining away train the next network iteration to recognize inanimate-but-occasionally-moved objects in still images, even when they not themselves being moved.

We train this architecture on optical flow from unlabeled real-world video datasets, producing a network that estimates high-quality Spelke-object segmentations on still images drawn from such videos.  We call the resulting network the Excitatory-Inhibitory Segment Extraction Network (EISEN). We show EISEN to be robust even when the objects and configurations in the training videos and test images are distinct. In what follows, we review the literature on related works, describe the EISEN architecture and training methods in detail, show results on both complex synthetic datasets and real-world videos, and analyze algorithmic properties and ablations.

%% file: Figures/fig1.tex
\begin{figure}[t]
\centering
\includegraphics[width=0.95\textwidth]{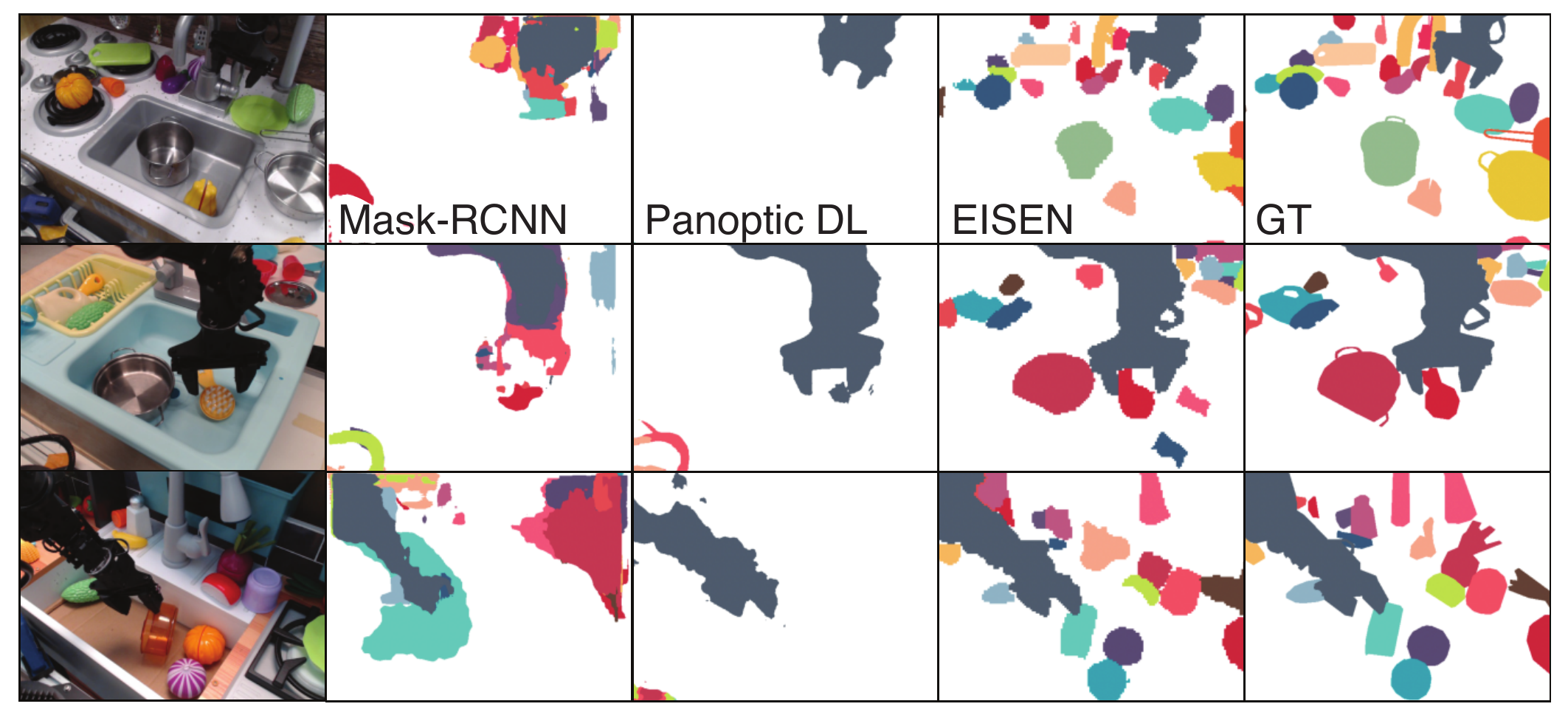}
\caption{\small \textbf{Unsupervised Segmentation of Spelke Objects.} Two standard object segmentation architectures, Mask-RCNN and Panoptic DeepLab, largely fail to learn to detect Spelke objects in the \textbf{Bridge} dataset without dense, categorical supervision. In contrast, our approach (EISEN) can detect these objects, without any supervision, via motion-based bootstrapping: learning to predict what moves together, then using top-down inference to segregate arm from object motion.}
\label{fig:headliner}
\vspace{-0.3in}
\end{figure}

%% file: Figures/fig2.tex
\begin{figure}[htb]
\centering
\includegraphics[width=0.95\textwidth]{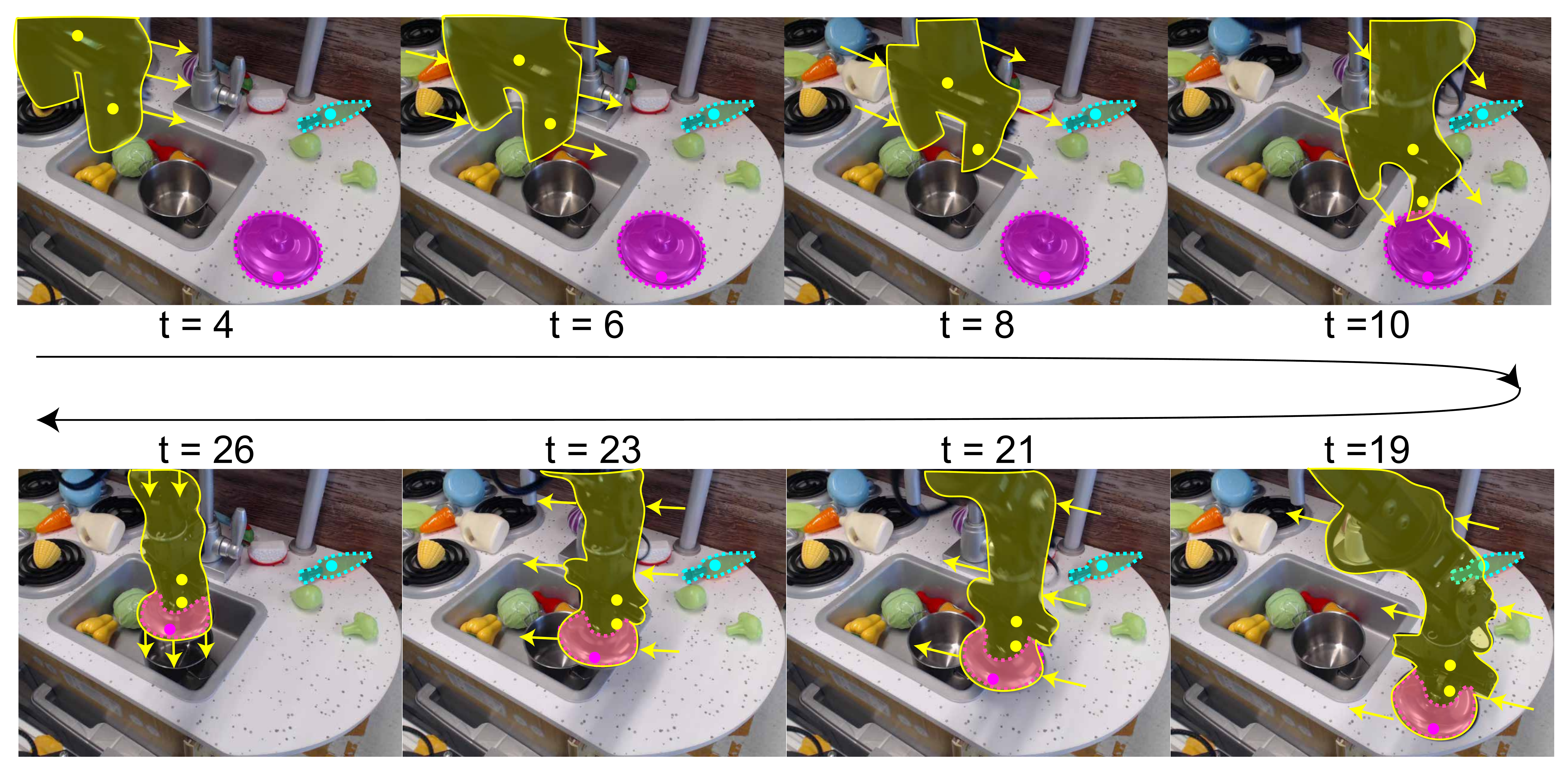}
\caption{\small \textbf{Two Challenges of Learning Spelke Objects.} \textbf{(Top Row)} Motion in real-world video is sparse. Thus, pairwise inferences about whether two points are moving together (e.g. the yellow points on the robot arm) or are not moving together (e.g. any yellow-cyan pairing) are valid.  However, pointwise motion-based inferences of whether a point is in a Spelke object or not will have many false negatives (e.g. points in the cyan object). \textbf{(Bottom row)} Inanimate objects (e.g. magenta lid) only move when moved by something else (e.g. the robotic arm), requiring explaining away of the apparent motion correlation (e.g. yellow-magenta pairs in the bottom row).} 
\label{fig:challenges}
\vspace{-0.3in}
\end{figure}

%% file: Sections/related_work_CR.tex
\section{Related Work}
\textbf{Segmentation as bottom-up perceptual grouping.}
The Gestalt psychologists discovered principles according to which humans group together elements of a scene, such as feature similarity, boundary closure, and correlated motion (``common fate'')~\cite{todorovic2008gestalt}.
This inspired classical computer vision efforts to solve segmentation as a bottom-up graph clustering problem~\cite{shi2000normalized,peng2013survey}.
Although these approaches achieved partial success, they have proved difficult to adapt to the enormous variety of objects encountered in real-world scenes like robotics environments.
Thus, today's most successful algorithms instead aim to segment objects by learning category-specific cues on large, labeled datasets~\cite{he2017mask,cheng2020panoptic,zhu2020deformable}.

\textbf{Unsupervised and category-agnostic segmentation.}
Several recent approaches have tried to dispense with supervision by drawing on advances in self-supervised object \textit{categorization}.
DINO, LOST, and Token-Cut perform ``object discovery'' by manipulating the attention maps of self-supervised Vision Transformers, which can be considered as maps of affinity between an image patch and the rest of the scene~\cite{caron2021emerging,simeoni2021localizing,wang2022self}.
PiCIE learns to group pixels in an unsupervised way by encouraging particular invariances and equivariances across image transformations.
While these early results are encouraging, they apply more naturally to \textit{semantic} segmentation than to grouping individual Spelke objects (instance segmentation): to date, they are mostly limited either to detecting a single object per image or to grouping together all the objects of each category.
The GLOM proposal~\cite{hinton2021represent} sketches out an unsupervised approach for constructing ``islands'' of features to represent object parts or wholes, which is similar to our grouping mechanism; but it does not provide a specific algorithmic implementation.
We find the architectural particulars of EISEN are essential for successful object segmentation in real-world images (see \textbf{Ablations}).

\textbf{Object discovery from motion.}
A number of unsupervised object discovery methods can segment relatively simple synthetic objects but struggle on realistic scenes~\cite{greff2019multi,locatello2020object,du2020unsupervised,kabra2021simone}.
When applied to the task of \textit{video object segmentation}, Slot Attention-based architectures can segment realistic moving objects~\cite{kipf2021conditional,yang2021self}, but none of these methods uses motion to learn to segment the majority of objects that are static at any given time.
Several approaches discover objects via motion signals, making a similar argument to ours for motion revealing physical structure~\cite{sabour2021unsupervised,du2020unsupervised,arora2021learning,tangemann2021unsupervised,tsao2021topological,dorfman2013learning,ross2008segmentation}.
However, they have been limited to segmenting a narrow range of objects or scenes.

We hypothesize that generalization to realistic, complex scenes benefits greatly from affinity-based grouping and learning.
In this respect, our work is heavily inspired by PSGNet, an unsupervised affinity-based network that learns to segment scenes from both object motion and other grouping principles~\cite{bear2020learning}.
We make two critical advances on that work: (1) replacing its problematic (and non-differentiable) Label Propagation algorithm with a neural network; 
and (2) introducing a bootstrapping procedure that uses top-down inference to explain raw motion observations in terms of confidently grouped objects.
In combination, these novel contributions allow EISEN to accurately perform a challenging task: the static segmentation of real-world objects without supervision.

%% file: Sections/methods_CR.tex
\input{Figures/fig3}

\section{Methods}

\subsection{The EISEN Architecture}
EISEN performs \textit{unsupervised, category-agnostic segmentation of static scenes}:
it takes in a single $H \times W \times 3$ RGB image and outputs a segmentation map of shape $H' \times W'$.
EISEN and baseline models are trained on the optical flow predictions of a RAFT network~\cite{teed2020raft} pretrained on Sintel~\cite{Butler:ECCV:2012}.
RAFT takes in a pair of frames, so EISEN requires videos for training but not inference.

\textbf{Overall concept.} The basic idea behind EISEN is to construct a high-dimensional feature representation of a scene (of shape $H' \times W' \times Q$) that is almost trivial to segment. 
In this desired representation, all the feature vectors $\mathbf{q}_{ij}$ that belong to the same object are aligned (i.e., have cosine similarity $\approx 1$) and all feature vectors that belong to distinct objects are nearly orthogonal (cosine similarity $\approx 0$). 
A spatial slice of this feature map looks like a set of flat object segment ``plateaus,'' so we call it the \textit{plateau map} representation.
Object segments can be extracted from a plateau map by finding clusters of vectors pointing in similar directions.

The plateau map is inherently relational: both building and extracting segments from it are straightforward given accurate pairwise affinities between scene elements. 
EISEN therefore consists of three modules applied sequentially to a convolutional feature extractor backbone (Figure \ref{fig:bootstrapping}):
\begin{enumerate}
    \item \textit{Affinity Prediction}, which computes pairwise affinities between features;
    \item \textit{Kaleidoscopic Propagation} (KProp), a graph RNN that aligns the vectors of a plateau map by passing messages on the extracted affinity graph;
    \item \textit{Competition}, an RNN that imposes winner-take-all dynamics on the plateau map to extract object segments and suppress redundant activity.
\end{enumerate}
All three modules are differentiable, but only Affinity Prediction has trainable parameters. 
We use the ResNet50-DeepLab backbone in Panoptic-DeepLab\cite{cheng2020panoptic}, which produces output features of shape $H/4\times W/4 \times 128$. 

\textbf{Affinity Prediction}. 
This module computes affinities $A(i,j,i',j')$ between pairs of extracted feature vectors $\mathbf{f}_{ij}, \mathbf{f}_{i'j'}$.
Each feature vector is embedded in $\mathbb{R}^D$ with linear key and query functions, and the affinities are given by standard softmax self-attention plus row-wise normalization:
\begin{align}
    &\tilde{A}^{ij}_{i'j'} = \text{Softmax} \left (\frac{1}{\sqrt{D}}(W_{k}\mathbf{f}_{ij})(W_q\mathbf{f}_{i'j'})^T \right),
    &A^{ij}_{i'j'} = \tilde{A}^{ij}_{i'j'} \ /\  \max_{\{i',j'\}}\tilde{A}^{ij}_{i'j'}.
\end{align}
To save memory, we typically compute affinities only within a $25\times25$ grid around each feature vector plus a random sample of long-range ``global'' affinities.

\input{Figures/fig4}
\textbf{Kaleidoscopic Propagation}.
The KProp graph RNN (Figure \ref{fig:kprop_comp}) is a smooth relaxation of the discrete Label Propagation (LProp) algorithm~\cite{gregory2010finding}.
Besides being nondifferentiable, LProp suffers from a ``label clashing'' problem: once a cluster forms, the discreteness of labels makes it hard for another cluster to merge with it.
This is pernicious when applied to image graphs, as the equilibrium clusters are more like superpixels than object masks~\cite{bear2020learning}.
KProp is adapted to the specific demands of image segmentation through the following changes:
\begin{itemize}
    \item Instead of integers, each node is labeled with a continuous vector $\mathbf{q}_{ij} \in \mathbb{R}^Q$; the full hidden state at iteration $s$ is $h_{s} \in \mathbb{R}^{N \times Q}$.
    \item The nondifferentiable message passing in LProp is replaced with two smooth operations: each node sends (1) \textit{excitatory} messages to its high-affinity neighbors, which encourages groups of connected nodes to align; and (2) \textit{inhibitory} messages to its low-affinity neighbors, which orthogonalizes disconnected node pairs. These messages cause clusters of nodes to merge, split, and shift in a pattern reminiscent of a kaleidoscope, giving the algorithm its name.
    \item At each iteration, node vectors are rectified and $\ell^2$ normalized. Although softmax normalization produces (soft) one-hot labels, it reinstates ``label clashing'' by making the $Q$ plateau map channels compete. $\ell^2$ normalization instead allows connected nodes to converge on an intermediate value.
\end{itemize}
During propagation, the affinity matrix is broken into two matrices, $A^{+}, A^{-}$, for excitatory and inhibitory message passing, respectively. 
These are simply the original affinity matrix with all values above (resp., below) $0.5$ set to zero, then normalized by the sum of each row.
The plateau map $h_{0}$ is randomly initialized and for each of $S$ iterations is updated by
\begin{flalign}
    &h^{+}_{s} = h_{s} + A^{+}h_{s}, \\
    &h^{-}_{s} = h^{+}_s - A^{-}h^{+}_{s}, \\
    &h_{s+1} = \text{Norm}(\text{ReLu}(h^{-}_{s})),
\end{flalign}
where $\text{Norm}$ does $\ell^2$ normalization.
We find that convergence is faster if only one random node passes messages at the first iteration.

\textbf{Competition.}
Vector clusters in the final plateau map are generic points on the $(Q-1)$-sphere, not the (soft) one-hot labels desired of a segmentation map. 
The Competition module resolves this by identifying well-formed clusters, converting them to discrete \textit{object nodes}, and suppressing redundant activity (Figure \ref{fig:kprop_comp} bottom.)
First, $K$ \textit{object pointers} $\{p^k\} \in \mathbb{R}^2$ are randomly placed at $(h,w)$ locations on the plateau map and assigned \textit{object vectors} $\mathbf{p}^k \in \mathbb{R}^Q$ according to their positions;
an \textit{object segment} $m^k \in \mathbb{R}^{H \times W}$ for each vector is then given by its cosine similarity with the full plateau map:
\begin{align}
    &p^{k} = (p^k_h, p^k_w), &  
    &\mathbf{p}^k = h_{S}(p^k_h,p^k_w),
    &m^k = h_{S} \cdot \mathbf{p}^k.
\end{align}
Some of the masks may overlap, and some regions of the plateau map may not be covered by any mask.
We use recurrent winner-take-all dynamics to select a minimal set of object nodes that fully explains the map.
Let $\mathcal{J}(\cdot, \cdot)$ denote the Jaccard index and let $\theta$ be a threshold hyperparameter (set at 0.2 in all our experiments). 
Competition occurs between each pair of object vectors with masks satisfying $\mathcal{J}(m^k, m^{k'}) > \theta$;
the winner is the vector with greater total mask weight $\sum_{i,j}m^k$.
An object that wins every pairwise competition is \textit{retained}, while all others are \textit{deactivated} by setting their masks to zero (Figure \ref{fig:kprop_comp} bottom.)
This process is repeated for a total of $R$ iterations by re-initializing each deactivated object $(p^l, \mathbf{p}^l, m^l = 0)$ on parts of the plateau map that remain uncovered, $U = 1 - \sum_k m^k$.
Thus the Competition module retains a set of $M <= K$ nonzero (soft) masks, which are then softmax-normalized along the $M$ dimension to convert them into a one-hot pixelwise segmentation of the scene.

\input{Sections/bootstrapping_CR}

%% file: Figures/fig3.tex
\begin{figure}[htb]

\centering
\includegraphics[width=0.95\textwidth]{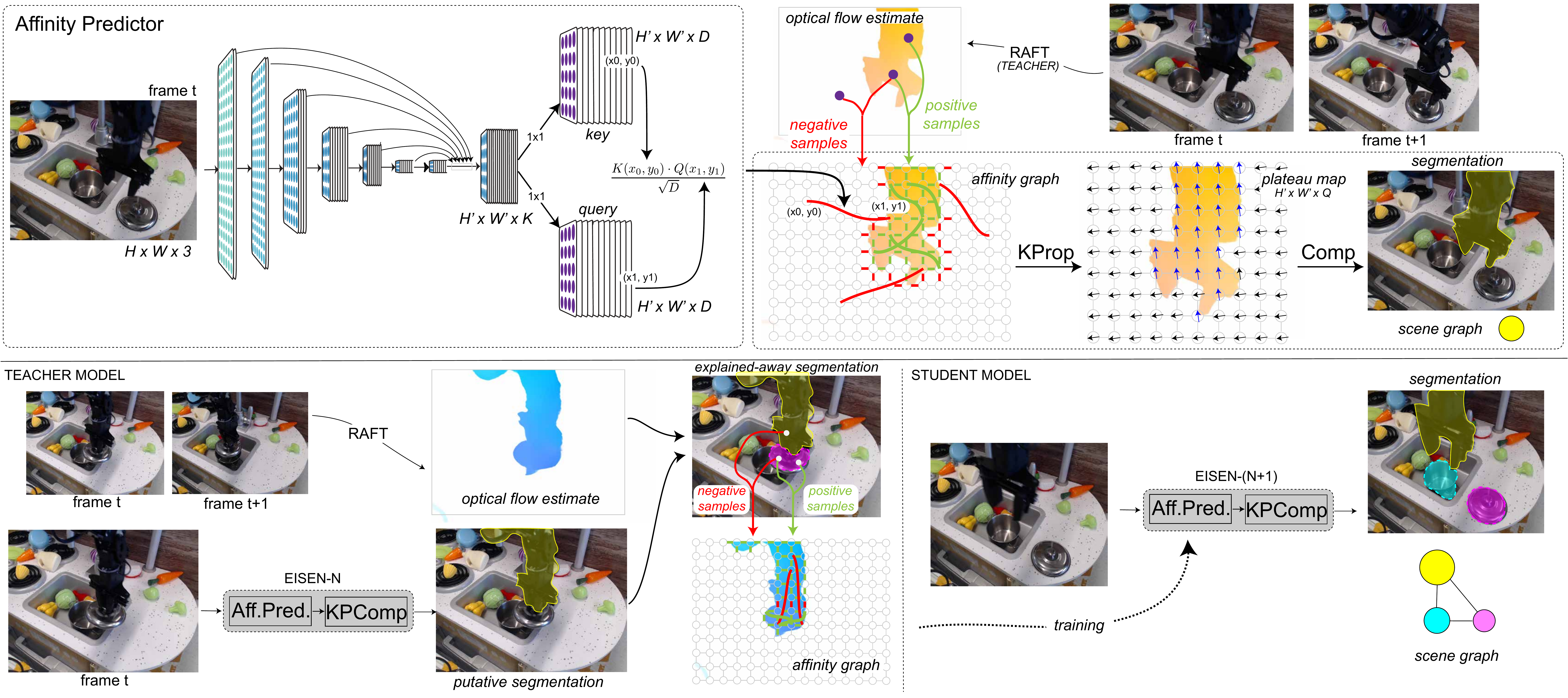}
\caption{\small \textbf{The EISEN architecture and training process.} \textbf{(Top: Architecture)} The EISEN architecture consists of (i) an Affinity Predictor module which extracts a pairwise affinity graph for each scene, and (ii) the KProp-Competition module, which converts the affinity graph into an actual segmentation. The affinity predictor is trained to predict thresholded optical flow estimates computed via the RAFT algorithm, with positive samples corresponding to pairs of moving points (green affinity graph edges), and negative samples corresponding to moving-nonmoving point pairs (red edges). Edges are computed for all pairs of close-by points and a sampling of further-separated point pairs.  Segments are extracted from the affinity graph via a two-stage mechanism consisting of Kaleidoscopic Propagation and inter-node Competition (see text and Fig. \ref{fig:kprop_comp} for more details). \textbf{(Bottom: Iterative Training)} Differences between RAFT optical flow estimates and high-confidence segments from static stage-$N$ EISEN outputs are ``explained away'' by positing the existence of new Spelke objects, which are then used to supervised the stage-$(N+1)$ EISEN model.}
\label{fig:bootstrapping}
\end{figure}

%% file: Figures/fig4.tex
\begin{figure}[!ht]
\centering
\includegraphics[width=0.95\textwidth]{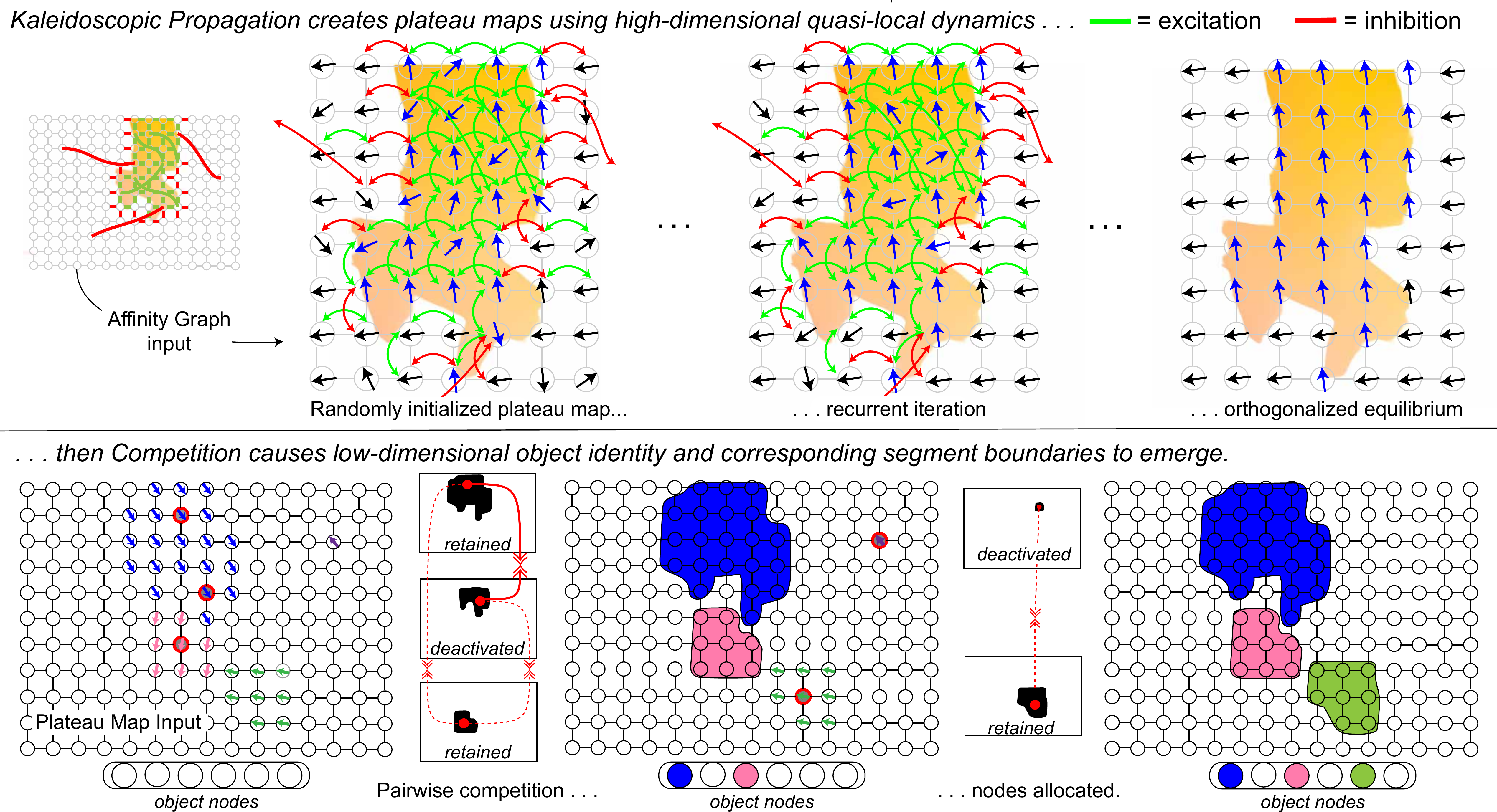}
\caption{\small \textbf{Kaleidoscopic Propagation and Competition.} \textbf{(Top Row: KProp)} For each node in an input affinity graph, a random normalized $Q$-dimensional vector is allocated (blue and black arrows).  The KProp module is a graph RNN that implements high-dimensional quasi-local dynamics, with  affinities $A$ corresponding to excitatory connections and inverted affinities $1-A$ corresponding to inhibitory connections. These dynamics are repeated a fixed number iterations, quickly coming to equilibrium at a ``plateau map'' in which candidate segments correspond to nearly-orthogonal domains in the $Q$-dimensional vector field each of which is nearly-flat. \textbf{(Bottom Row: Competition)} Plateau maps are converted into segmentations by having ``object nodes'' compete for ownership of points within the plateau map. A set of putative object nodes are initialized with randomly located basepoints (red highlighted nodes). Each object node corresponds to an object mask consisting of plateau map locations with high $Q$-vector correlation to the vector at the basepoint. Pairs of object nodes with overlapping masks compete, with the overall-more-aligned node winning and suppressing alternates. Reinitialization then occurs only over non-covered territory. After a small number of iterations, the process equilibrates with the masks containing segment estimates, and the object nodes describing the scene graph.}
\label{fig:kprop_comp}
\vspace*{-0.6cm}
\end{figure}

%% file: Sections/bootstrapping_CR.tex
\subsection{Training EISEN via Spelke Object Inference}
Because KProp and Competition have no trainable parameters, training EISEN is tantamount to training the affinity matrix $A$.
This is done with a single loss function: the row-wise KL divergence between $A$ and a \textit{target connectivity matrix}, $\mathcal{C}$, restricted to the node pairs determined by \textit{loss mask}, $\mathcal{D}$: 
\begin{align}
    \mathcal{L}_{\text{EISEN}} = \sum_{i,j}\text{KLDiv}(\mathcal{D}^{ij}_{i'j'} \odot A^{ij}_{i'j'}, \mathcal{D}^{ij}_{i'j} \odot \mathcal{C}^{ij}_{i'j'}).
\end{align}
To compute $\mathcal{C}$ and $\mathcal{D}$ we consider pairs of scene elements $(a,b)$ that project into image coordinates $(i,j)$ and $(i',j')$, respectively.
If only one element of the pair is moving (over long enough time scales), it is likely the two elements do not belong to the same Spelke object;
when neither is moving, there is no information about their connectivity, so no loss should be computed on this pair.
This is the core physical logic -- ``Spelke object inference'' -- by which we train EISEN.

\textbf{Computing connectivity targets from motion.}
Let $\mathcal{I}(\cdot)$ be a motion indicator function, here $\mathcal{I}(a) = (|\textbf{flow}_{ij}| > 0)$, where $\textbf{flow}$ is a map of optical flow.
The logic above dictates setting
\begin{align}
    &\mathcal{C}^{ij}_{i'j'} \leftarrow\ 0\ \textbf{if}\ (\mathcal{I}(a)\ \textbf{xor}\ \mathcal{I}(b)), \\
    &\mathcal{D}^{ij}_{i'j'} \leftarrow\ 1\ \textbf{if}\ (\mathcal{I}(a)\ \textbf{or}\ \mathcal{I}(b))\ \textbf{else}\ 0.
\end{align}
To learn accurate affinities there must also pairs with $C^{ij}_{i'j'} = 1$ that indicate when two scene elements belong to the same object.\footnote{
If scenes are assumed to have at most one independent motion source, these are simply the pairs with $\mathcal{I}(a) == \mathcal{I}(b) == 1$. This often holds in robotics scenes (and is perhaps the norm in a baby's early visual experience) but not in many standard datasets (e.g. busy street scenes.) We therefore handle the more general case.}
When a scene contains multiple uncorrelated motion sources, the optical flow map has an appearance similar to a plateau map (e.g. Figure \ref{fig:davis_motion}, second column.)
This allows the flow map to be segmented into multiple motion sources as if it \textit{were} a plateau map using the Competition algorithm (see Supplement for details.)
The positive pairs in the connectivity target can then be set according to
\begin{align}
    &\tilde{\mathcal{C}}^{ij}_{i'j'} \leftarrow 1\ \textbf{if}\ (\mathcal{S}_{M}(a) == \mathcal{S}_{M}(b))\ \textbf{and} (\mathcal{I}(a) == \mathcal{I}(b) == 1)\ \textbf{else}\ 0,
\label{eqn:mseg}
\end{align}
where $\mathcal{S}_{M}$ is the estimated map of motion segments. Any elements of the background are assumed to be static with $\mathcal{S}_{M}(a) == \mathcal{I}(a) == 0$ (see Supplement.)

\textbf{Segmenting correlated motion sources by top-down inference.}
Naïve application of Equation (\ref{eqn:mseg}) cannot handle the case of an agent moving a Spelke object (as in Figure \ref{fig:challenges}) because agent and object will be moving in concert and thus will appear as a single ``flow plateau.''
However, a \textit{non-naïve observer} might have already seen the agent alone moving and have learned to segment it via static cues.
If this were so, the agent's pixels could be ``explained away'' from the raw motion signal, isolating the Spelke object as its own target segment (Figure \ref{fig:bootstrapping}, lower panels.)
Concretely, let $\mathcal{S}_{\mathcal{T}}$ be a map of \textit{confidently segmented objects} output by a teacher model, $\mathcal{T}$ (see Supplement for how EISEN computes confident segments.)
Any scene elements that do not project to confident segments have $\mathcal{S}_{\mathcal{T}}(a) = 0$.
Then the final loss mask is modified to include all pairs with at least one moving \textbf{or} confidently segmented scene element,
\begin{align}
    &\hat{\mathcal{D}}^{ij}_{i'j'} \leftarrow 1\ \textbf{if}\ ((\mathcal{S}_M(a) + \mathcal{S}_{\mathcal{T}}(a) > 0)\ \textbf{or}\ (\mathcal{S}_M(b) + \mathcal{S}_{\mathcal{T}}(b) > 0))\ \textbf{else}\ 0.
\end{align}
Explaining away is performed by overwriting pairs in Equation (\ref{eqn:mseg}) according to whether two scene elements belong to the same or different confident segments, \textit{regardless of whether they belong to the same motion segment}:
\begin{align}
    &\hat{\mathcal{C}}^{ij}_{i'j'} \leftarrow (\mathcal{S}_{\mathcal{T}}(a) == \mathcal{S}_{\mathcal{T}}(b))\ \textbf{if}\ (\mathcal{S}_{\mathcal{T}}(a) + \mathcal{S}_{\mathcal{T}}(b) > 0)\ \textbf{else}\ \tilde{\mathcal{C}}^{ij}_{i'j'}.
\label{eqn:cfinal}
\end{align}
Thus the final connectivity target, $\hat{\mathcal{C}}$, combines Spelke object inference with the confident teacher predictions, defaulting to the latter in case of conflict.

\textbf{Bootstrapping.}
Since objects that appear moving more often should be confidently segmented earlier in training, it is natural to \textit{bootstrap}, using one (frozen) EISEN model as teacher for another student EISEN (Figure \ref{fig:bootstrapping}.)
After some amount of training, the student is frozen and becomes the teacher for the next round, as a new student is initialized with the final weights of the prior round.
Although bootstrapping could be continued indefinitely, we find that EISEN confidently segments the majority of Spelke objects after three rounds.

%% file: Sections/results_CR.tex
\section{Results}
\input{Tables/table1_tdw_val}
\subsection{Datasets, Training, and Evaluation}
Full details of datasets, training, and evaluation are in the Supplement.
Briefly, we train EISEN and baseline models on motion signals from three datasets:
\textbf{Playroom}, a ThreeDWorld~\cite{gan2020threedworld} dataset of realistically simulated and rendered objects (2000 total) that are invisibly pushed;
the \textbf{DAVIS2016}~\cite{perazzi2016benchmark} video object segmentation dataset, which we repurpose to test \textit{static} segmentation learning in the presence of background motion;
and \textbf{Bridge}~\cite{ebert2021bridge}, a robotics dataset in which human-controlled robot arms move a variety of objects.

We compare EISEN to the (non-differentiable) affinity-based SSAP~\cite{gao2019ssap}, the Transformer-based DETR~\cite{carion2020end}, the centroid prediction-based Panoptic DeepLab (PDL)~\cite{cheng2020panoptic}, and the region proposal-based Mask-RCNN~\cite{he2017mask}.
All baselines require pixelwise segmentation supervision, for which we use the same motion signals as EISEN except for the conversion to pairwise connectivity.
Because they were not designed to handle sparse supervision, we tune baseline object proposal hyperparameters to maximize recall.
All models are evaluated on mIoU between ground truth and best-matched predicted segments; DETR and Mask-RCNN are not penalized for low precision.

\input{Figures/fig5}
\subsection{Learning to segment from sparse object motion}
\textbf{EISEN outperforms standard architectures at motion-based learning.}
Baseline segmentation architectures easily segment the \textbf{Playroom}-\textit{val} set when given full supervision of all objects (Table \ref{table:playroom_val}, Full supervision.)
When supervised only on RAFT-predicted optical flow, however, these models perform substantially worse (Table \ref{table:playroom_val}, Motion supervision) and exhibit characteristic qualitative failures (Figure \ref{fig:baselines_by_sup}), such as missing or lumping together objects.

EISEN, which treats object motion as an exclusively \textit{relational} learning signal, performs well whether given full or motion-only supervision (Table \ref{table:playroom_val}.)
Moreover, in contrast to the baselines, EISEN also accurately segments most objects in \textit{test} scenes that differ from its training distribution in background, object number, and multi-object occlusion patterns (Figure \ref{fig:baselines_by_sup}; see Supplement.)
These results suggest that only EISEN learns to detect the class of \textit{Spelke objects} -- the category-agnostic concept of ``physical stuff that moves around together.''
Interestingly, the strongest motion-supervised baseline is Mask-RCNN, which may \textit{implicitly} use relational cues in its region proposal and non-maximal suppression modules to partly exclude false negative static regions of the scene.

\subsection{Self-supervised segmentation of real-world scenes.}
\input{Figures/fig6}
\textbf{Learning to segment in the presence of background motion.}
The \textbf{Playroom} dataset has realistically complex Spelke objects but unrealistically simple motion.
In particular, its scenes lack background motion and do not show the agentic mover of an object.
Most video frames in the \textbf{DAVIS2016} dataset~\cite{perazzi2016benchmark} have both object and (camera-induced) background motion, so we use it to test whether a useful segmentation learning signal can be extracted and used to train EISEN in this setting.
Applying Competition to flow plateau maps often exposes a large background segment, which can be suppressed to yield a target object motion segment (Figure \ref{fig:davis_generalization}A; also see Supplement.)
When this motion signal is used to train EISEN, the \textit{static} segmentation performance on \textit{held-out scenes} is 0.52, demonstrating that motion-based self-supervision supports learning of complex Spelke objects real scenes (Figure \ref{fig:davis_generalization}B.)

\input{Tables/table4_realworld_complex}

\textbf{Unsupervised segmentation of the Bridge robotics dataset.}
We train EISEN for three rounds of bootstrapping to segment Spelke objects in \textbf{Bridge} (see Methods). 
EISEN's segmentation of \textbf{Bridge} scenes dramatically improves with each round (Table \ref{table:bridge_val} and Figure \ref{fig:bridge_rounds}).
In the first round, the model mainly learns to segment the robot arm, which is expected because this object is seen moving more than any other and the untrained EISEN teacher outputs few confident segments that could overwrite the raw motion training signal.
In the subsequent rounds, top-down inference from the pretrained EISEN teacher modifies the raw motion signal; the improvement during these rounds suggests that top-down inference about physical scene structure can extract better learning signals than what is available from the raw image or motion alone.
In contrast to EISEN, neither Mask-RCNN nor Panoptic DeepLab segment most of the objects either after applying the same bootstrapping procedure or when pretrained on COCO with categorical supervision (Table \ref{table:bridge_val} and Figure \ref{fig:headliner}.)
EISEN's combination of bottom-up grouping with top-down inference thus enables unsupervised segmentation of Spelke objects in real scenes.

\input{Figures/fig7}
\subsection{Ablations of the EISEN architecture}
\input{Tables/table2_ablations}
\input{Tables/table5_dino}
\textbf{Ablating KProp and Competition.}
EISEN performance on \textbf{Playroom} is nearly equal when using all affinity pairs versus using local and a small sample of long-range pairs ($<7\%$ of total), though it drops slightly if long-range pairs are omitted (Table \ref{table:ablations}).
This suggests that plateau map alignment is mainly a local phenomenon and that grouping with EISEN relies heavily on local cues.

In contrast, the architectural components of KProp and Competition are essential for EISEN's function.
When either excitatory or inhibitory messages are ablated, or when using Softmax rather than $\ell^2$-normalization, performance drops nearly to zero (Table \ref{table:ablations}.)
Moreover, the Competition module is better at extracting segments from the final plateau map than simply taking the Argmax over the channel dimension;
this is expected, since the $\ell^2$-normalization in KProp does not encourage plateau map clusters to be one-hot vectors.

KProp and Competition are both RNNs, so their function may change with the number of iterations.
Performance saturates only with $>30$ KProp iterations and drops to near zero with a single iteration, implying that sustained message passing is essential: EISEN cannot operate as a feedforward model.
In contrast, Competition requires only a single iteration on \textbf{Playroom} data (Table \ref{table:ablations}).

\textbf{Ablating Affinity Prediction.}
Finally, we compare EISEN's affinities to other affinity-like model representations.
Object segments can be extracted from the attention maps of a self-supervised Vision Transformer (DINO~\cite{caron2021emerging}) using KProp and Competition, but their accuracy is well below EISEN's;
prior graph-based segmentation methods~\cite{shi2000normalized,gregory2010finding,frey2007clustering} do not detect \textbf{Playroom} objects as well as KProp-Competition (Table \ref{table:dino}; see Supplement.)
These experiments imply that EISEN is a better source of affinities than the (statically trained) DINO attention maps and that EISEN's grouping network best makes use of both sources.

%% file: Tables/table1_tdw_val.tex

\begin{table}[t]
\centering
\setlength{\tabcolsep}{4pt}
\caption{\small \textbf{Performance (mIoU) of instance segmentation models on the TDW-Playroom dataset}. Models with \textbf{full} supervision receive masks for all movable objects in the scene at training time; models with \textbf{motion} supervision receive the optical flow predicted by RAFT.}
\label{table:playroom_val}
\begin{tabular}{lccccc}
\toprule
\multirow{2}{*}{Model} 
      & \multicolumn{2}{c}{Full supervision} 
      &
          & \multicolumn{2}{c}{Motion supervision} \\
\cmidrule{2-3}\cmidrule{5-6}
  & val & test & & val & test  \\  
  \midrule
  SSAP & 0.802 & 0.575 & & 0.295 & 0.235 \\
  DETR & 0.860 & 0.647 & & 0.297 & 0.258 \\
  Panoptic DeepLab & \textbf{0.870} & 0.608 & &  0.620 & 0.373\\
  Mask-RCNN  & 0.713 & 0.387 & & 0.629 & 0.467\\
\midrule
  EISEN & 0.788 & \textbf{0.675} & & \textbf{0.730} & \textbf{0.638}\\
\bottomrule
\end{tabular}
\end{table}

%% file: Figures/fig5.tex
\begin{figure}[htb]
\centering
\includegraphics[width=0.95\textwidth]{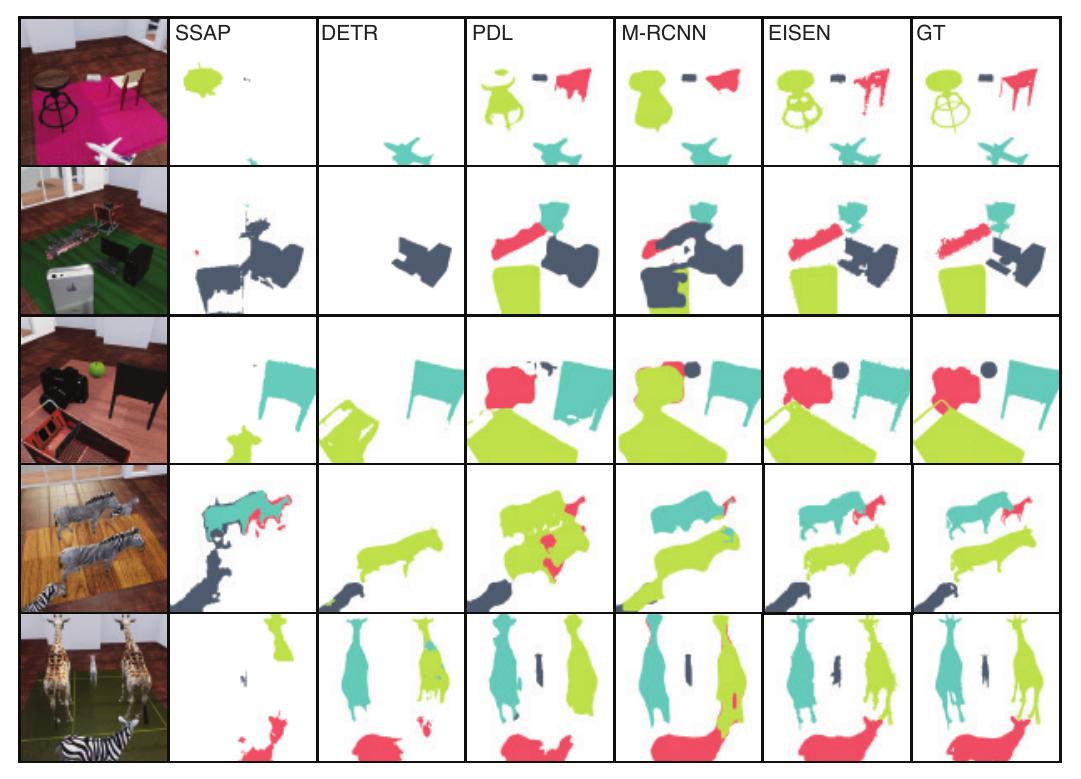}
\caption{\textbf{EISEN outperforms baselines at learning to segment from motion.} Segmentation predictions of EISEN and each baseline are shown for examples from the \textbf{Playroom} \textit{val} set (top three rows) and \textit{test} set (bottom two rows.) Baselines frequently lump, miss, and distort the shapes of objects. EISEN is able to capture fine details (e.g. the chair and giraffe legs) and segment closely spaced objects of similar appearance, (e.g. the zebras.)} 
\label{fig:baselines_by_sup}
\vspace*{-0.5cm}
\end{figure}

%% file: Figures/fig6.tex
\begin{figure}
\centering
\includegraphics[width=0.95\textwidth]{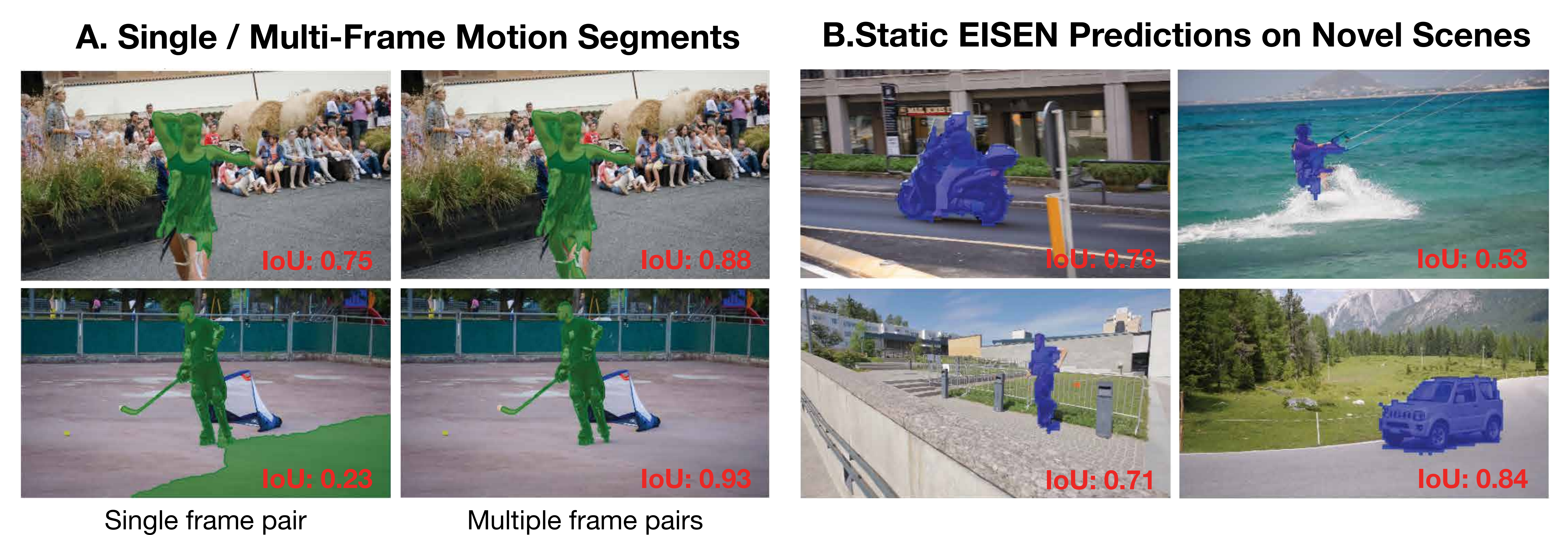}
\caption{\small \textbf{EISEN learns to segment objects in \textit{static, held-out scenes} on DAVIS2016.} (A) Confident \textit{teacher} segments computed from multiple frame pairs are better than those from a single frame pair. (B) Without any motion information, EISEN segments objects in \textit{single RGB images} from held-out scenes.}  
\label{fig:davis_generalization}
\vspace*{-\baselineskip}
\end{figure}

%% file: Tables/table4_realworld_complex.tex
\setlength{\tabcolsep}{4pt}

\begin{table}[htbp]

\centering
\caption{\small \textbf{Performance on Bridge after each round of bootstrapping.} EISEN improves at segmentation across three rounds by using its own inference pass to create better supervision signals. Neither Mask-RCNN nor Panoptic DeepLab perform well whether bootstrapped or pretrained on COCO.}
\label{table:bridge_val}
\begin{tabular}{lcccc}
\toprule
Model
& \multicolumn{1}{p{2.cm}}{\centering Round 1}
& \multicolumn{1}{p{2.cm}}{\centering Round 2}
& \multicolumn{1}{p{2.cm}}{\centering Round 3}
& \multicolumn{1}{p{2.cm}}{\centering Pretrained}\\
\midrule
MaskRCNN & 0.053 & 0.081 & 0.102 & 0.070 \\
Panoptic DeepLab & 0.051 & 0.056 & 0.057 & 0.175 \\
EISEN & 0.336 & 0.453 & 0.551 & -\\
\bottomrule
\end{tabular}
\end{table}

\setlength{\tabcolsep}{1.4pt}

%% file: Figures/fig7.tex
\begin{figure}
\centering
\includegraphics[width=0.95\textwidth]{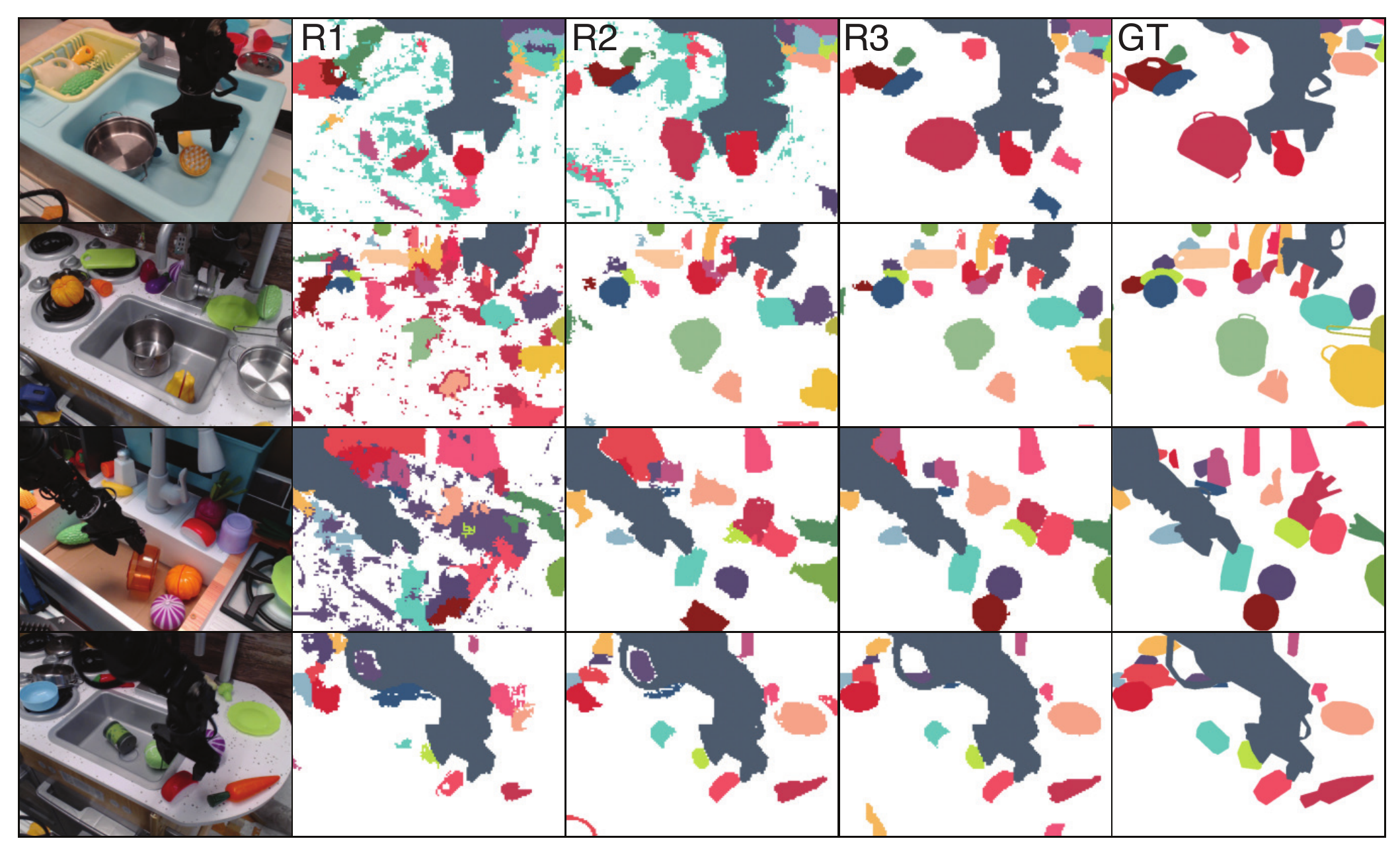}
\caption{\small \textbf{EISEN improves at segmenting Spelke objects with each round of bootstrapping.} After the first round of bootstrapping (R1), EISEN can segment the arm but few other objects well. Subsequent rounds (R2 and R3) substantially improve both the number of objects detected and their segmentation quality.}
\label{fig:bridge_rounds}
\vspace*{-0.5cm}
\end{figure}

%% file: Tables/table2_ablations.tex
\begin{table}[t]
\centering
\caption{\small \textbf{Ablations of EISEN.} Altering the architectural components of KProp or Competition drastically degrades performance, but only a small sample of long-range affinities are necessary (Left). Lowering the number of RNN iterations for KProp or Comp gradually degrades performance (Right).}
\label{table:ablations}
\begin{subtable}[h]{0.52\textwidth}
\centering
\begin{tabular}{cccccc}
\toprule
messages & affinity & norm & readout & mIoU \\
\midrule
  Ex+Inb & Loc+Glob & $\ell^2$ & Comp & 0.730 \\
  Ex+Inb & Loc & $\ell^2$ & Comp & 0.700 \\
  Ex+Inb & Full & $\ell^2$ & Comp & 0.732 \\
  Ex+Inb & Loc+Glob & $\ell^2$ & Argmax & 0.676 \\
  Ex & Loc+Glob & $\ell^2$ &  Comp & 0.036 \\
  Inb & Loc+Glob & $\ell^2$ &  Comp & 0.036 \\
  Ex+Inb & Loc+Glob & softmax & Comp & 0.036\\
\bottomrule
\end{tabular}
  \end{subtable}
  \hfill 
\begin{subtable}[h]{0.44\textwidth}
\centering
\setlength{\tabcolsep}{4pt}
\begin{tabular}{cccccc}
\toprule
KProp iters~& Comp iters~& mIoU \\
\midrule
  40 & 3 & 0.730 \\
  30 & 3 & 0.720 \\
  20 & 3 & 0.697 \\
  10 & 3 & 0.389 \\
  1 & 3 & 0.052 \\
  40 & 2 & 0.730\\
  40 & 1 & 0.729 \\
\bottomrule
\end{tabular}
  \end{subtable}
\end{table}
\setlength{\tabcolsep}{1.4pt}

%% file: Tables/table5_dino.tex
\begin{table}[t]
\centering
\caption{\textbf{Comparison of DINO and EISEN affinities with different graph clustering algorithms.} EISEN affinities are downsampled to the same size as DINO affinities for a fair comparison}
\label{table:dino}
\setlength{\tabcolsep}{3mm}
\begin{tabular}{cccccc}
\toprule
  Model & Spectral clustering & LabelProp & AffinityProp & KProp+comp \\
\midrule
  DINO & 0.354 & 0.135 & 0.255 & 0.545 \\
  EISEN & 0.062 & 0.084 &  0.319 &  0.684 \\
\bottomrule
\end{tabular}

\end{table}
\setlength{\tabcolsep}{1.4pt}

%% file: Sections/conclusion.tex
\section{Conclusion}
We have proposed EISEN, a fully differentiable, graph-based grouping architecture for learning to segment Spelke objects.
While our algorithm performs on par with prior segmentation models when fully supervised (Table \ref{table:playroom_val}), its main strength is an ability to learn \textit{without supervision}:
by applying top-down inference with its own segmentation predictions, it progressively improves motion-based training signals.
These key architecture and learning innovations are critical for dealing with the challenges of unsupervised, category-agnostic object segmentation in real-world scenes (Figure \ref{fig:challenges}.) 
Since EISEN is based on the principle of grouping things that move together, it cannot necessarily address higher-level notions of ``objectness'' that include things rarely seen moving (e.g., houses and street signs.)
It will therefore be important in future work to explore the relationship between motion-based and motion-independent object learning and identify deeper principles of grouping that extend to both.

\textbf{Acknowledgements}
J.B.T is supported by NSF Science Technology Center Award CCF-1231216.
D.L.K.Y is supported by the NSF (RI 1703161 and CAREER Award 1844724) and hardware donations from the NVIDIA Corporation.
J.B.T. and D.L.K.Y. are supported by the DARPA Machine Common Sense program.
J.W. is in part supported by Stanford HAI, Samsung, ADI, Salesforce, Bosch, and Meta. 
D.M.B. is supported by a Wu Tsai Interdisciplinary Scholarship and is a Biogen Fellow of the Life Sciences Research Foundation. 
We thank Chaofei Fan and Drew Linsley for early discussions about EISEN.

%% file: Sections/supplement.tex
\makeatletter 
\renewcommand{\thefigure}{S\@arabic\c@figure}
\renewcommand{\thetable}{S\arabic{table}}
\makeatother
\setcounter{figure}{0}
\setcounter{table}{0}

\title{Supplemental Material for ``Unsupervised Segmentation in Real-World Images via Spelke Object Inference''}

\authorrunning{Chen et al.}
\titlerunning{Unsupervised Segmentation via Spelke Object Inference}

\author{
Honglin Chen \inst{1},
Rahul Venkatesh \inst{1},
Yoni Friedman \inst{4},
Jiajun Wu \inst{1}, \\
Joshua B. Tenenbaum \inst{4},
Daniel L. K. Yamins \inst{1,2,3}{**},
Daniel M. Bear \inst{2,3}{**}}

\institute{Department of Computer Science, Stanford \and 
Department of Psychology, Stanford \and
Wu~Tsai~Neurosciences~Institute, Stanford \and
Department of Brain and Cognitive Sciences and CBMM, MIT}

\footnotetext[1]{$^{**}=$ Equal senior authorship}
\maketitle
\section{Additional Methods}
\input{Sections/supplement_text/eisen_details}
\input{Sections/supplement_text/dataset_details}
\input{Sections/supplement_text/architecture_details}
\input{Sections/supplement_text/training_details}
\input{Sections/supplement_text/evaluation_details}
\newpage
\section{Additional Visualization}
\input{Figures/figS2}
\input{Figures/figS3}

%% file: Sections/supplement_text/eisen_details.tex
\subsection{Details and Extensions of EISEN}
\textbf{Learning a prior over Spelke objects.}
Relational supervision is natural for motion-based learning in part because motion is sparse.
But the output segments of a well-trained EISEN are \textit{not} sparse, so they can be used for learning \textit{non}-relational features of Spelke objects.
For instance, an image patch on the nearer side of a depth edge may ``look like'' it lies on the interior of an object segment -- a cue known as \textit{border ownership}.
EISEN can take advantage of these non-relational features by learning a nonrandom, pixelwise initialization for KProp.

Specifically, we let the $Q$ channels of the plateau map code for possible object centroid locations.
Let $Q_H, Q_W$ be the encoding resolutions of height and width, with $Q_H \cdot Q_W = Q$. 
Then the \textit{centroid encoding} is constructed by creating the $Q_H \times Q_W \times Q$ feature tensor $q$ defined by
\begin{align}
    q_{ijk} = 1\ \textbf{if}\ (k\ ==\ iQ_W + j)\ \textbf{else}\ 0,
\end{align}
then bilinearly upsampling this tensor from size $(Q_H,Q_W,Q)$ to the usual plateau map resolution $(H',W',Q) = (H/4,W/4,Q)$ and $\ell^2$ normalizing along the channel dimension.
In this encoding, there is not just a ground truth segmentation map but also a ground truth plateau map -- namely, the one in which each feature vector has a value equal to the encoded centroid of the true object segment it belongs to.

We train two modified RAFT networks\footnote{We simply replace the output head that predicts predicts a $H \times W \times 2$ flow map with one that predicts the $H \times W \times 1$ ``objectness'' logits or $H \times W \times 2$ centroid offsets.} to (1) classify whether each pixel belongs in an EISEN-predicted Spelke object or not and, if it belongs to an object, (2) predict the relative offset between that pixel's location and the centroid of the object segment it belongs to.
The plateau map initialization $h_0$ is then given by the ``objectness''-masked, predicted centroid encodings of each pixel.
In a loose analogy between KProp and Ising-like models of magnetic dipole dynamics, this initialization plays the role of a pulsed external field.

Interestingly, learning a KProp initialization does not improve quantitative results on the \textbf{Playroom} dataset, though in some cases it appears to help discover an object that is only partially segmented with random plateau map initialization (Figure \ref{fig:kp_prior}.)
It may be that the learned initialization is helpful in some ways but harmful in others, such as by degrading fine details.
It is notable that passing the learned initializations directly through Competition without running any iterations of KProp (Figure \ref{fig:kp_prior}, \textit{-KProp}) performs better than all baselines, achieving an mIoU of 0.660 on the \textbf{Playroom} \textit{val} set.
This suggests that EISEN can best take advantage of \textit{non-relational} cues to segment these scenes, but that there may be a low upper bound to performance when relational cues are not used.
\input{Figures/KP_prior}

\textbf{Constructing and segmenting a flow plateau map.}
The procedure for creating a flow plateau map to explain away background motion (see Figure \ref{fig:davis_motion}) is similar to constructing the centroid encoding described above.
Given a $H \times W \times 2$ optical flow map $\mathcal{F}$, we linearly normalize all flow values $(\mathcal{F}_x, \mathcal{F}_y)$ to the range $[-1,1]\times[-1,1]$.
Then, each pixel's flow vector $\mathcal{F}_{ij}$ is embedded in a $Q$-dimensional space by constructing the $Q_H \times Q_W \times Q$ centroid encoding tensor $q$ as above, then bilinearly sampling from this encoding with the normalized flow values,
\begin{align}
    \tilde{\mathcal{F}_{ij}} = q(\mathcal{F}^{\text{norm}}_x, \mathcal{F}^{\text{norm}}_y),
\end{align}
where $\tilde{\mathcal{F}}$ is the $Q$-channel ``flow plateau map'' and $q(i',j')$ denotes ``soft'' indexing with normalized image coordinates (i.e., bilinear sampling, as opposed to hard indexing, $q_{i'j'}$.) 

Competition is run on the flow plateau map with $K = 32$ maximum segments and $R = 3$ rounds to detect the motion segments $\mathcal{S}_M$.
The largest of these by area is assumed to be the background; the motion indicator tensor $\mathcal{I}$ is created by taking the complement of this background segment, and its segment identity in $\mathcal{S}_M$ is set to $0$.
Examples of motion segments computed on \textbf{DAVIS2016} are shown in Figure \ref{fig:davis_motion} (third column.)
When these segments are used as self-supervision for EISEN, the resulting \textit{static segments} (Figure \ref{fig:davis_motion}, fourth column) can sometimes be more accurate than the motion segments, likely because affinities computed from single-frame appearance cues (such as color or texture similarity) generalize better than motion similarity, which varies substantially from one frame pair to another.
\input{Figures/DAVIS_motion}

\textbf{Computing confident segments.}
To isolate the confident segment predictions of an EISEN teacher model, we take advantage of the random initialization of the plateau map input to KProp: confident segments are those that are consistent across inference runs with different initializations.
Specifically, If $\{\mathcal{S}_{\mathcal{T}}^l |\ l = 1,2,...,L\}$ are the segments output by $L = 5$ runs of the teacher model, then we compute a set of ``meta-affinities'' based on how often two scene elements belong to the same segment:
\begin{align}
    &\hat{A}(a,b) = \frac{1}{L}\sum_{l}(\mathcal{S}_{\mathcal{T}}^l(a) == \mathcal{S}_{\mathcal{T}}^l(b)).
\end{align}
The meta-affinities are then converted to confident segments by applying KProp and Competition, keeping only the largest connected component of each predicted segment, and removing all segments of area $<10$ pixels. 
This has the effect of filtering out low-confidence segments because, in practice, KProp only yields well-formed pixel clusters when the input (meta-)affinities are highly confident, i.e. close to $0$ or $1$.
In the initial round of training, for which the EISEN teacher has not been pretrained, there are few if any confident segments, and the training target mainly reduces to Equation (\ref{eqn:mseg}).
Thus, successive rounds of bootstrapping tend to have more accurate training targets (such as by separating Spelke objects from agents and by providing pseudolabels for static objects) even though all rounds use \textit{the same rule} for inferring the connectivity target and loss mask.

%% file: Figures/KP_prior.tex
\begin{figure}
\centering
\includegraphics[width=0.95\textwidth]{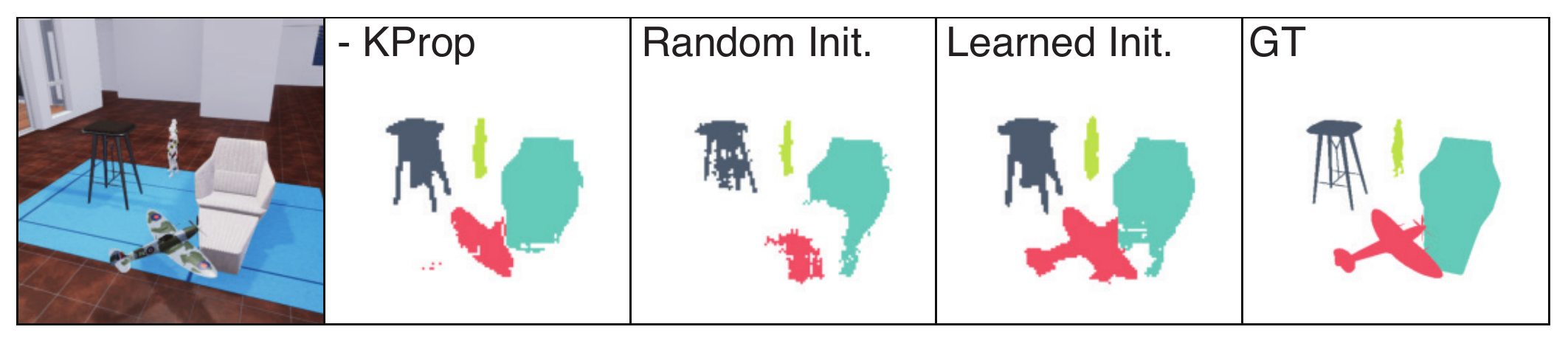}
\caption{\small \textbf{Learning a Spelke object prior for KProp.}} 
\label{fig:kp_prior}
\end{figure}

%% file: Figures/DAVIS_motion.tex
\begin{figure}
\centering
\includegraphics[width=0.95\textwidth]{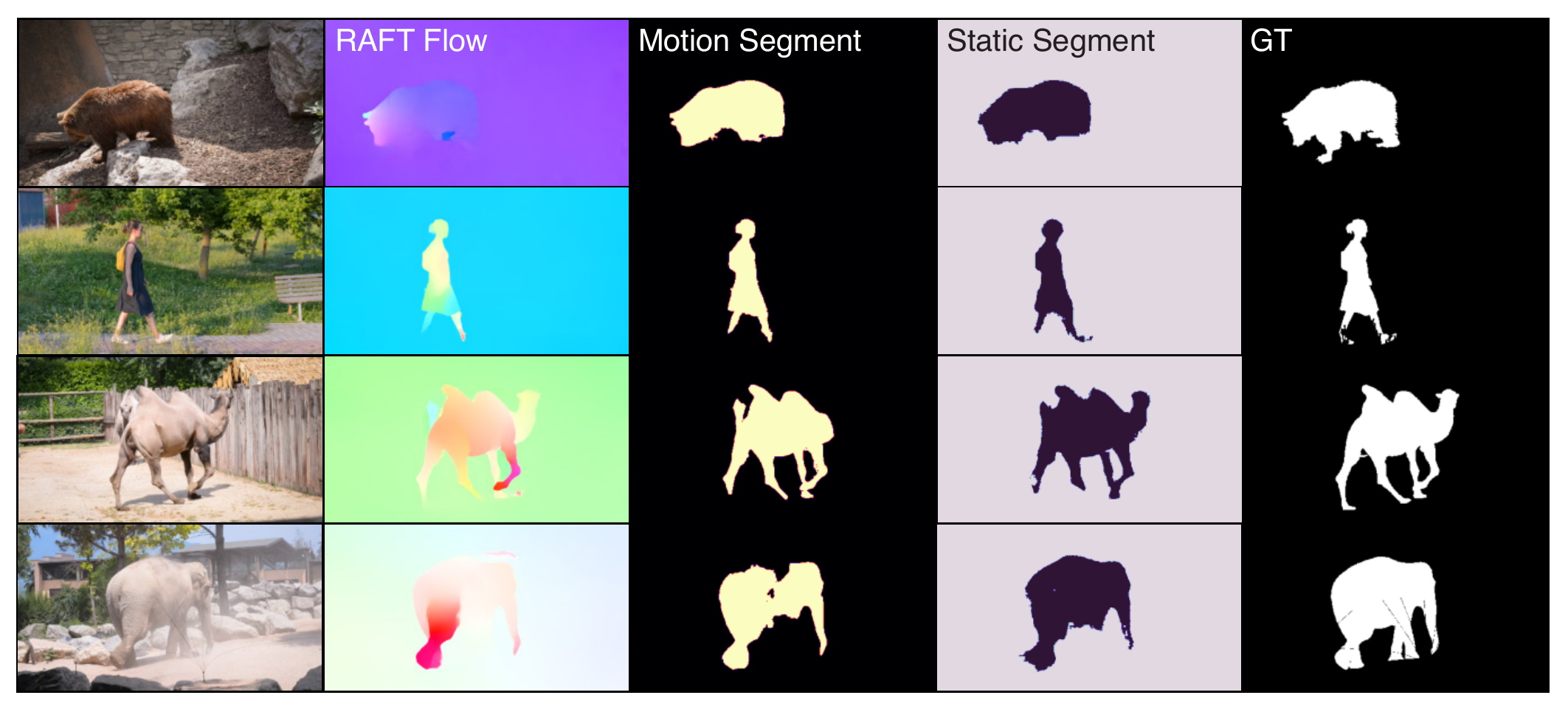}
\caption{\small \textbf{Explaining away background motion with Competition.} RAFT predictions cannot be threshholded directly. However, applying Competition to the flows, as though they were plateau maps, isolates the background and yields viable supervision targets. Training a \textit{static} segmentation model with these targets can pick up details that the motion segments miss.} 
\label{fig:davis_motion}
\vspace*{-\baselineskip}
\end{figure}

%% file: Sections/supplement_text/dataset_details.tex
\subsection{Datasets}
\textbf{Playroom.} 
The \textbf{Playroom} dataset was generated with ThreeDWorld~\cite{gan2020threedworld} using custom code, which will be made public.
The dataset consists of 40000 videos. Each video shows four objects placed on an immobile, randomly colored and textured ``rug'' in a tiled room. 
The objects are drawn from a pool of 2000 models and scaled so that they fit within the room. The camera is randomly positioned and pointed so that at least three of the objects are within view. At the fifth frame of each video, an invisible force is applied to one of the objects that pushes it toward another object; the scene ends when the pushed object comes to rest or leaves the field of view. Within a given video, only the pushed object is able to move.

Each object model is seen moving in $40000 / 2000 = 20$ videos.
We hold out 4000 videos, use 500 of these as the \textit{val} set, and train on the remaining 36000.
EISEN and all baselines are trained only on the fifth frame of each video, with the supervising RAFT flow computed between the fifth and sixth frames.

In addition to the \textbf{Playroom} \textit{train} and \textit{val} datasets, we also generated a \textit{test} dataset of $30$ scenes that departs from the model training set in several ways.
Specifically, it contains scenes with multiple copies of a particular object (e.g. the giraffes and zebras in the bottom two rows of Figure \ref{fig:baselines_by_sup}), scenes set in a different room (such that the background is different), and scenes with simply textured ``primitive'' objects containing, occluding, and colliding with each other; these primitive objects are not seen moving in the training set.
Thus, the \textbf{Playroom} \textit{test} set measures how well segmentation models generalize to new object arrangements and contexts.
Both the \textit{test} set and its generation script will be released along with all code.

\textbf{Bridge.} 
The Bridge dataset consists of 7200 demonstrations for 71 kitchen-themed tasks collected in 10 different environments~\cite{ebert2021bridge}. Each demonstration shows a robotic arm executing a semantically meaningful task (e.g. put spoon into pot) in a household kitchen environment with different robotic positions, background, and lighting conditions. Each demonstration is collected with 3-5 camera viewpoints concurrently. 7 out of 10 environments were collected at the University of California, Berkeley. The three remaining environments were collected at the University of Pennsylvania. We train and evaluate models on the subset of the Bridge dataset collected at the University of California, Berkeley. In particular, we train on a total of 5881 randomly selected demonstrations. Since ground-truth segmentation annotations are not provided for the Bridge dataset, we manually annotate 50 held-out images for evaluating the validation performance.

%% file: Sections/supplement_text/architecture_details.tex
\subsection{Model Architecture Details}
\textbf{EISEN backbone.}
EISEN uses the ResNet50-DeepLab convolutional network as its feature extractor~\cite{lin2017feature}. To ensure that EISEN is trained in an unsupervised manner, we randomly initialize the backbone parameters using He initialization \cite{he2015delving}, instead of using a ImageNet-pretrained backbone. The backbone is trained end-to-end along with the Affinity Prediction module.

\textbf{EISEN input and output resolution.}
We use whole images as inputs without applying data augmentation. The input resolution is 512$\times$512, 270$\times$480, and 480$\times$640 for the \textbf{Playroom}, \textbf{DAVIS}, and \textbf{Bridge} datasets respectively. The backbone outputs feature tensors at 1/4 of the input resolution, and EISEN predicts the affinities and segmentation masks at the output resolution of the backbone. 
Output segments are upsampled to the original resolution for evaluation.

\textbf{EISEN hyperparameters.}
For all experiments on \textbf{Playroom} and \textbf{DAVIS}, we set the Affinity Prediction key and query dimension $D = 32$, the plateau map dimension $Q = 256$, and the maximum number of objects detectable by Competition $K = 32$. For \textbf{Bridge}, which contains more objects per scene, we increase to $K = 256$. 
By default we run KProp for $S = 40$ iterations and Competition for $R = 3$ rounds.

\textbf{Baselines.}
The original baseline models are trained in a category-specific way with a separate semantic head for predicting object categories. However, given the absence of semantic supervision in a category-agnostic setting, we convert the semantic heads to binary objectness classifiers. In particular, we change the semantic loss function from the multi-class cross entropy to the binary cross entropy, which encourages the semantic head to predict 1 for Spelke objects and 0 otherwise. The semantic head architectures are identical to the original models, except for the output dimension in the final readout layer. 

\begin{table}
\centering
\caption{Comparison of backbones and parameter count}
\label{table:parameter_count}
\setlength{\tabcolsep}{3mm}
\begin{tabular}{llc}
\toprule
  Model & Backbone & Parameters \\
\midrule
  SSAP~\cite{gao2019ssap} & ResNet34-FPN & 48M  \\
  DETR~\cite{carion2020end} & ResNet50 & 41M \\
  MaskRCNN~\cite{he2017mask} & ResNet50-FPN & 43M \\
  Panoptic-Deeplab~\cite{cheng2020panoptic} & ResNet50-DeepLab & 30M \\
  \midrule
  EISEN & ResNet50-DeepLab & 40M\\
\bottomrule
\end{tabular}

\end{table}
\setlength{\tabcolsep}{1.4pt}

%% file: Sections/supplement_text/training_details.tex
\subsection{Model Training}
\textbf{EISEN training protocol.}
We adopt a similar training protocol in Panoptic-Deeplab\cite{cheng2020panoptic}.
In particular, we use the ‘poly’ learning rate policy \cite{liu2015parsenet} with an initial learning rate of 0.005, and optimize with Adam~\cite{luo2019adaptive} without weight decay. On the \textbf{Playroom} and \textbf{DAVIS2016} datasets, we train EISEN with a batch size of 8 for 200k iterations. On the \textbf{Bridge} dataset, we train EISEN with a batch size of 8 for 60k, 20k, 20k iterations for three rounds of bootstrapping, respectively.
Training EISEN for 100k iterations on 8 GPUs takes 20 hours.
Because \textbf{DAVIS2016} is not typically used to evaluate static segmentation (rather than video object segmentation and tracking), we developed a protocol in which 45 out of 50 scenes are used for (motion-based) training and 5 out of 50 are held-out and shown as \textit{static images only} to the pretrained EISEN model for testing.

\textbf{Baseline training protocol.}
For a fair comparison with EISEN, we train baselines with whole images as inputs and without applying data augmentation. The baseline models are trained from scratch without using ImageNet-pretrained weights. Other settings are the same as the original MaskRCNN\cite{he2017mask}, Panoptic-Deeplab\cite{cheng2020panoptic}, DETR\cite{carion2020end} and SSAP \cite{gao2019ssap} models. For evaluating the baseline models at inference time, we perform a grid search to find thresholds for pixelwise ``objectness'' classification that maximize mIoU on 500 images from the training set. Note that because of how object segment proposals are scored against ground truth segments, DETR and Mask-RCNN are not penalized for using a low objectness threshold to make many proposals.
For running multiple rounds of bootstrapping with Mask-RCNN and Panoptic DeepLab, we apply the same teacher-student setup as with EISEN.
Confident segments from baseline models are determined by taking all object proposals above their optimal cross-validated confidence thresholds (see Model Evaluation below.)

%% file: Sections/supplement_text/evaluation_details.tex
\subsection{Model Evaluation}
\textbf{EISEN inference time.}
Although EISEN contains two RNNs (KProp and Competition) that may be unrolled for many iterations, its inference time is not substantially longer than that of baselines:
EISEN takes 155ms to perform inference on a single 512 x 512 image with 30 iterations of KProp and 3 iterations of Competition, compared to 65 ms for Mask-RCNN.
Because KProp iterations are implemented as sparse matrix multiplications, unrolling this RNN for many iterations is not particularly slow.
Note also that during training, $\mathcal{L}_{EISEN}$ is applied directly to the \textit{affinities} $A^{ij}_{i'j'}$, such that it is not necessary to perform expensive backpropagation-through-time on the KProp RNN. 
(KProp and Competition do need to be run to compute teacher object segments for bootstrapping, but no gradients need to be computed from the teacher model.)

\textbf{Matched mIoU.} Our metric for how well a model segments a scene's Spelke objects is the intersection over union (IoU) between predicted and ground truth segments, averaged over ground truth segments in each image, and then averaged across images in the evaluation dataset.

The mIoU for a given image is computed by finding the best one-to-one match between predicted and ground truth segments using linear sum assignment; a single predicted segment therefore cannot match to multiple ground truth segments.
Because EISEN outputs a ``panoptic'' instance segmentation map (i.e. every pixel is assigned to exactly one segment), there is no ambiguity about which predicted segment should be matched with the ground truth.
For baselines that output overlapping object segment \textit{proposals} (in this work, DETR~\cite{zhu2020deformable} and Mask-RCNN~\cite{he2017mask}), we compute a pseudo-panoptic segmentation map by assigning each pixel that falls into \textit{any} predicted segment to the highest confidence prediction. 
This ensures fair comparison to EISEN and other panoptic segmentation models (like SSAP~\cite{gao2019ssap} and Panoptic DeepLab~\cite{cheng2020panoptic}) that cannot benefit from making multiple segment proposals at each spatial location.
We think that this segmentation metric is the one most appropriate to our goal of parsing scenes into Spelke objects, since an agent that wanted to \textit{use} an object-centric scene representation would ultimately need to choose which single segmentation proposal to act on at any given time.

\textbf{Computing an affinity map from Vision Transformers.}
To convert DINO~\cite{caron2021emerging} or other Vision Transformer attention maps to an affinity-like output, we use the \texttt{vit\_small} architecture with a patch size of $\textrm{8x8}$. We compute the attention map using the final self-attention layer.
The affinity between two patches $p_1$ and $p_2$ is obtained by computing the normalized dot product between their respective query vectors, $q_1^h$ and $q_2^h$ for a given head $h$. Since Vision Transformers outputs have multiple attention heads, we use the average of the attention values computed across different heads,
\begin{align}
    \textit{Affinity} \hspace{0.5mm} (p_1, p_2) = \left( \underset{h}{\sum} \dfrac{q_1^h \cdot q_2^h }{|q_1^h| |q_2^h|}\right)/N_{heads},
\end{align}
where $N_{heads}$ is the number of attention heads.

%% file: Figures/figS2.tex
\begin{figure}[!ht]
\centering
\includegraphics[width=0.95\textwidth]{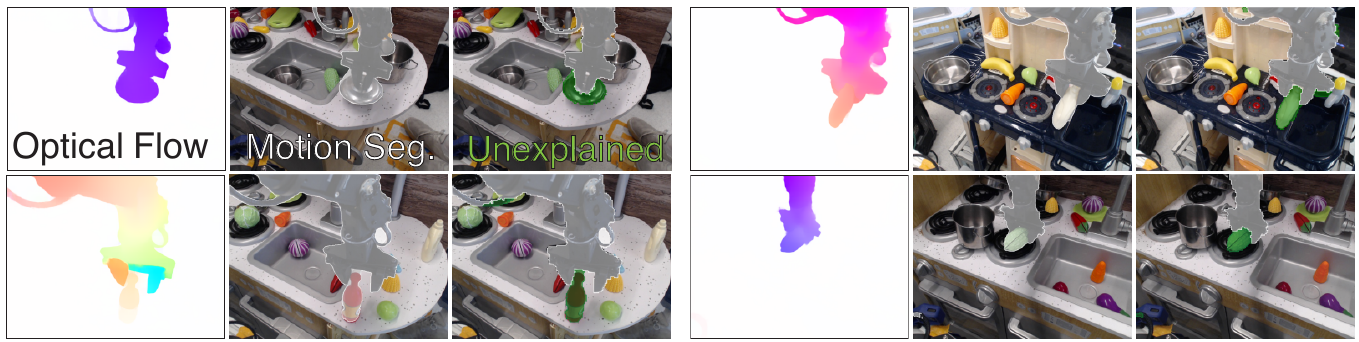}
\caption{\small \textbf{Explaining Away improves the motion supervision signal on Bridge.} Four training examples from the \textbf{Bridge} dataset showing the \textbf{Optical Flow} predicted by a pretrained RAFT model (left images); the \textbf{Motion Segment} yielded by simply thresholding the optical flow, indicated by a white overlay (center images); and the \textbf{Unexplained Motion} yielded by the \textbf{Explaining Away} bootstrapping process, indicated by a green overlay (right images). The motion explained away as the moving agent remains as the white overlay in the right images.}
\label{fig:explained_away}
\vspace*{-0.6cm}
\end{figure}

%% file: Figures/figS3.tex
\begin{figure}[!ht]
\centering
\includegraphics[width=0.95\textwidth]{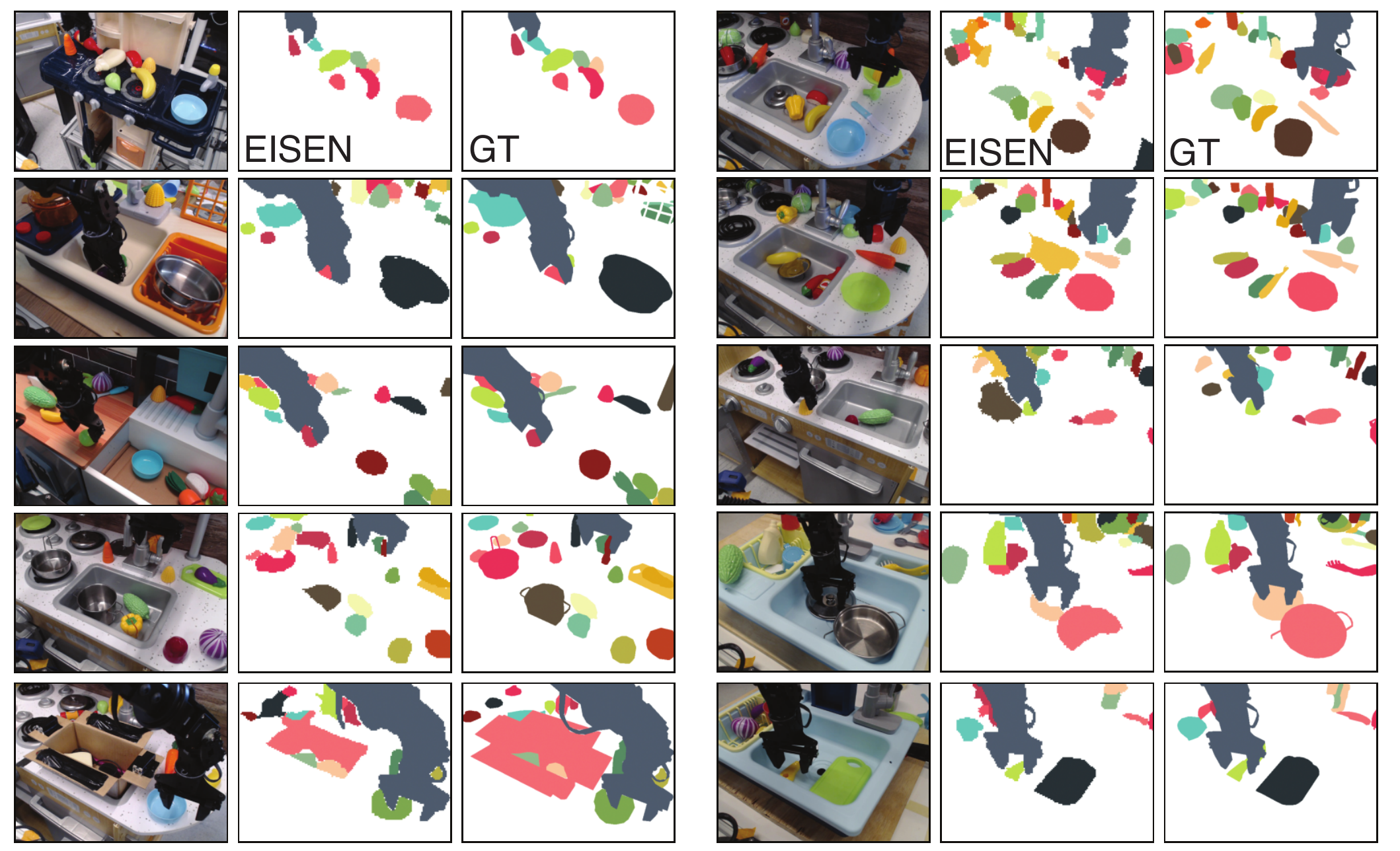}
\caption{\small \textbf{More examples of EISEN predictions on the Bridge dataset.}}
\label{fig:more_bridge}
\vspace*{-0.6cm}
\end{figure}